\documentclass[sigconf]{acmart}
\usepackage{graphicx} 
\usepackage{wrapfig}  
\usepackage[utf8]{inputenc}
\usepackage{tikz}
\usepackage{pgfplots}
\pgfplotsset{compat=newest}
\usepgfplotslibrary{groupplots}  
\usetikzlibrary{patterns}
\usepackage{booktabs}
\usepackage{pifont}
\usepackage{enumitem}
\usepackage{makecell}   
\usepackage{hyperref}
\usepackage{url}
\usepackage{multirow}
\usepackage{subcaption}
\AtBeginDocument{%
  }

\copyrightyear{2025}
\acmYear{2025}
\setcopyright{acmlicensed}
\acmConference[KDD '25]{Proceedings of the 31st ACM SIGKDD Conference on Knowledge Discovery and Data Mining V.2}{August 3--7, 2025}{Toronto, ON, Canada}
\acmBooktitle{Proceedings of the 31st ACM SIGKDD Conference on Knowledge Discovery and Data Mining V.2 (KDD '25), August 3--7, 2025, Toronto, ON, Canada}
\acmDOI{10.1145/3711896.3737429}
\acmISBN{979-8-4007-1454-2/2025/08}

\begin{document}

\title{TrustGLM: Evaluating the Robustness of GraphLLMs Against Prompt, Text, and Structure Attacks
}

\author{Qihai Zhang}
\authornote{Equal contribution}
\email{qz2086@nyu.edu}
\affiliation{%
  \institution{New York University} 
  \department{Center for Data Science}
  \city{New York}
  \country{USA}
}
\author{Xinyue Sheng}
\authornotemark[1]
\email{xs2334@nyu.edu}
\affiliation{%
  \institution{New York University Shanghai}
  \department{Data Science}
  \city{Shanghai}
  \country{China}
}

\author{Yuanfu Sun}
\email{yuanfu.sun@nyu.edu}
\affiliation{%
  \institution{New York University} 
  \department{Courant Institute}
  \city{New York}
  \country{USA}
}

\author{Qiaoyu Tan}
\authornote{Corresponding author}
\email{qiaoyu.tan@nyu.edu}
\affiliation{%
  \institution{New York University Shanghai}
  \department{Computer Science}
  \city{Shanghai}
  \country{China}
}

\renewcommand{\shortauthors}{Qihai Zhang, Xinyue Sheng, Yuanfu Sun, \& Qiaoyu Tan}

\begin{abstract}
Inspired by the success of large language models (LLMs), there is a significant research shift from traditional graph learning methods to LLM-based graph frameworks, formally known as GraphLLMs. GraphLLMs leverage the reasoning power of LLMs by integrating three key components: the textual attributes of input nodes, the structural information of node neighborhoods, and task-specific prompts that guide decision-making. Despite their promise, the robustness of GraphLLMs against adversarial perturbations remains largely unexplored—a critical concern for deploying these models in high-stakes scenarios. To bridge the gap, we introduce TrustGLM, a comprehensive study evaluating the vulnerability of GraphLLMs to adversarial attacks across three dimensions: text, graph structure, and prompt manipulations. We implement state-of-the-art attack algorithms from each perspective to rigorously assess model resilience. Through extensive experiments on six benchmark datasets from diverse domains, our findings reveal that GraphLLMs are highly susceptible to text attacks that merely replace a few semantically similar words in a node's textual attribute. We also find that standard graph structure attack methods can significantly degrade model performance, while random shuffling of the candidate label set in prompt templates leads to substantial performance drops. Beyond characterizing these vulnerabilities, we investigate defense techniques tailored to each attack vector through data-augmented training and adversarial training, which show promising potential to enhance the robustness of GraphLLMs. We hope that our open-sourced library will facilitate rapid, equitable evaluation and inspire further innovative research in this field. The benchmark code can be found in \url{https://github.com/Palasonic5/TrustGLM.git}.


\end{abstract}




\received{24 February 2025}

\maketitle

\section{Introduction}
Text-attributed Graphs (TAGs)—where nodes, edges, or both carry textual information—are becoming increasingly prominent in a wide range of applications such as social media analysis, e-commerce, and biomedical research. By integrating text with graph structures, TAGs provide richer contextual cues and deeper semantic relationships than traditional graphs. This inherent combination of relational and textual information has shown substantial promise in capturing complex interactions and improving downstream tasks ranging from recommendation systems~\cite{wang2019kgat, liu2021contextualized,fang2025uniglm} to information retrieval~\cite{wang2018information, shi2024retrieval,dietz2018utilizing}. As such, methods that effectively combine structural connectivity with textual content have garnered significant attention in both academia and industry.

A common approach to modeling TAGs leverages Graph Neural Networks (GNNs)~\cite{kipf2016semi, hamilton2017inductive}, which extend neural network operations to graph-structured data through message passing and aggregation. When applied to text graphs, GNNs typically fuse node features derived from textual embeddings with representations learned from the graph topology, enabling them to handle diverse graph tasks, including node classification, link prediction, and community detection. The wide adoption and success of GNNs have established them as a cornerstone technique in the field of TAGs research.

However, GNNs face critical limitations when handling text graphs. First, their reliance on local message passing hinders the capture of long-range dependencies~\cite{xu2018powerful, wu2020comprehensive}, which are often crucial in real-world applications where relevant information may reside in distant nodes. Second, by treating text as static node features, GNNs fall short in dynamically reasoning over rich semantic content, leading to inefficiencies in capturing linguistic nuances and domain-specific knowledge. Third, the dependence on graph structure makes GNNs vulnerable to noise and sparsity, resulting in reduced generalize when graph connectivity is incomplete or adversarially altered.

To address these challenges, a new paradigm-Graph Large Language Models (GraphLLMs) has recently emerged as a promising alternative~\cite{jin2024large,liu2024moleculargpt, li2023survey,fang2024gaugllm, ren2024survey}, integrating the powerful language understanding capabilities of Large Language Models (LLMs) into graph reasoning. Unlike traditional GNNs, GraphLLMs dynamically process textual attributes, capture complex semantic relationships, and incorporate graph structural information either by embedding structural cues into the LLM’s reasoning process~\cite{tang2024graphgpt} or by enhancing textual representations with graph context~\cite{chen2024llaga, sun2025graphicl, ye2024languagegraphneeds, wang2024llmszeroshotgraphlearners}. This approach overcomes GNNs’ limitations related to locality and static text modeling, offering a more expressive and adaptive framework for text graph learning.


Despite these advantages, GraphLLMs are highly vulnerable to adversarial manipulations, posing significant risks in high-stake applications. Specifically, vulnerabilities arise from three primary attack vectors: \textbf{A1: Graph Structure Attacks}, where adversaries alter the graph topology by inserting, deleting, or perturbing edges and nodes, thereby distorting relational dependencies and causing cascading failures in inference; \textbf{A2: Raw Text Attacks}, which modify the textual content associated with nodes or edges, leading to significant degradation in LLM-based reasoning since textual embeddings and model predictions are highly sensitive to even minor textual perturbations; and \textbf{A3: Prompt Manipulation Attacks}, in which attackers tamper with task-specific prompts or label sets, misleading the model into incorrect classifications or biased predictions, as GraphLLMs heavily depend on well-structured prompts to perform reasoning tasks effectively. Therefore, a natural question arises: \textit{How robust are GraphLLMs against structural, textual, and prompt-based adversarial attacks?} 

To answer this question, we present \textbf{TrustGLM}, a comprehensive benchmark for systematically evaluating the robustness of GraphLLMs against these adversarial attacks. Specifically, we implement \textbf{7} state-of-the-art attack algorithms across the three dimensions—structure, text, and prompt—to benchmark GraphLLMs' vulnerabilities and suggest effective defense techniques to enhance their robustness. Our key contributions are as follows: 

\begin{itemize}[noitemsep,leftmargin=*]
    \item \textbf{Systematic Robustness Evaluation:} We present TrustGLM, a novel benchmark that provides a standardized framework for assessing the robustness of GraphLLMs against structural, textual, and prompt-based adversarial attacks. The code and leaderboard will be publicly available and continuously updated to support ongoing research in secure and resilient GraphLLMs.
    \item \textbf{Novel Attack Paradigm:} We introduce a set of simple yet effective Prompt Manipulation Attacks, which disturb instruction prompts and label sets to examine the stability of GraphLLMs. Based on this, we establish a comprehensive taxonomy of GraphLLM attacks, providing a systematic framework for analyzing adversarial vulnerabilities.
    \item \textbf{Comprehensive Defense Implementation:} We implement targeted defense strategies—via data augmentation and adversarial training—to counteract each type of attack, thereby enhancing the robustness of GraphLLMs. Our study highlights critical security risks in GraphLLMs and lays the foundation for developing more resilient graph learning models on TAGs.
\end{itemize}


\section{Related Work}
Our work is closely related to the following two directions.

\textbf{LLM-based Graph Learning (GraphLLM).} This approach often has three key components. \textbf{(1) Graph Structure Understanding.} There are 2 ways to incorporate graph structure information in GraphLLM. (1) By using GNNs to generate node embeddings that capture both local and global relationships within the graph \cite{tang2024graphgpt, zhang2024graphtranslator, liu2024can, he2024unigraph} (2)By employing hop-based sampling methods to generate node sequences, ensuring that relevant graph neighborhoods are effectively represented for the LLM \cite{chen2024llaga, liu2024moleculargpt} \textbf{(2)Projector} is often used to align graph embeddings with the LLM embedding space, facilitating smooth integration between structured graph data and the LLM’s textual representation. \textbf{(3)Frozen pretrained LLMs} The language model itself remains unchanged while graph information is injected in a way that it can process naturally. This direct alignment between graph modality and pretrained LLM allows GraphLLMs to generalize across different graph reasoning tasks without additional training.  


\noindent \textbf{Adversarial Attacks.} Existing attack methods can be categorized by adversary knowledge. The categories are: black-box attack \cite{dai2018adversarial, sun2020adversarial}, gray-box attack \cite{zugner2018adversarial}, and white-box attack \cite{wang2021robustdeepneuralnetwork, zhang2019adversarialattacksdeeplearning, geisler2021prbcd, wang2020scalable}. In white-box attacks, the attacker has full knowledge of the model’s architecture, parameters, loss function, activation functions, and even training data, which often leads to highly effective attacks. Conversely, black-box attacks operate under the assumption that none of these internal details are available, and generate adversarial examples by querying the model and monitoring output changes on the test dataset. Gray-box attacks operate under partial knowledge, such as access to the training data but not model information, often relying on surrogate model strategies \cite{zugner2018adversarial}.

\begin{figure*}[htbp]
    \includegraphics[width=\linewidth]{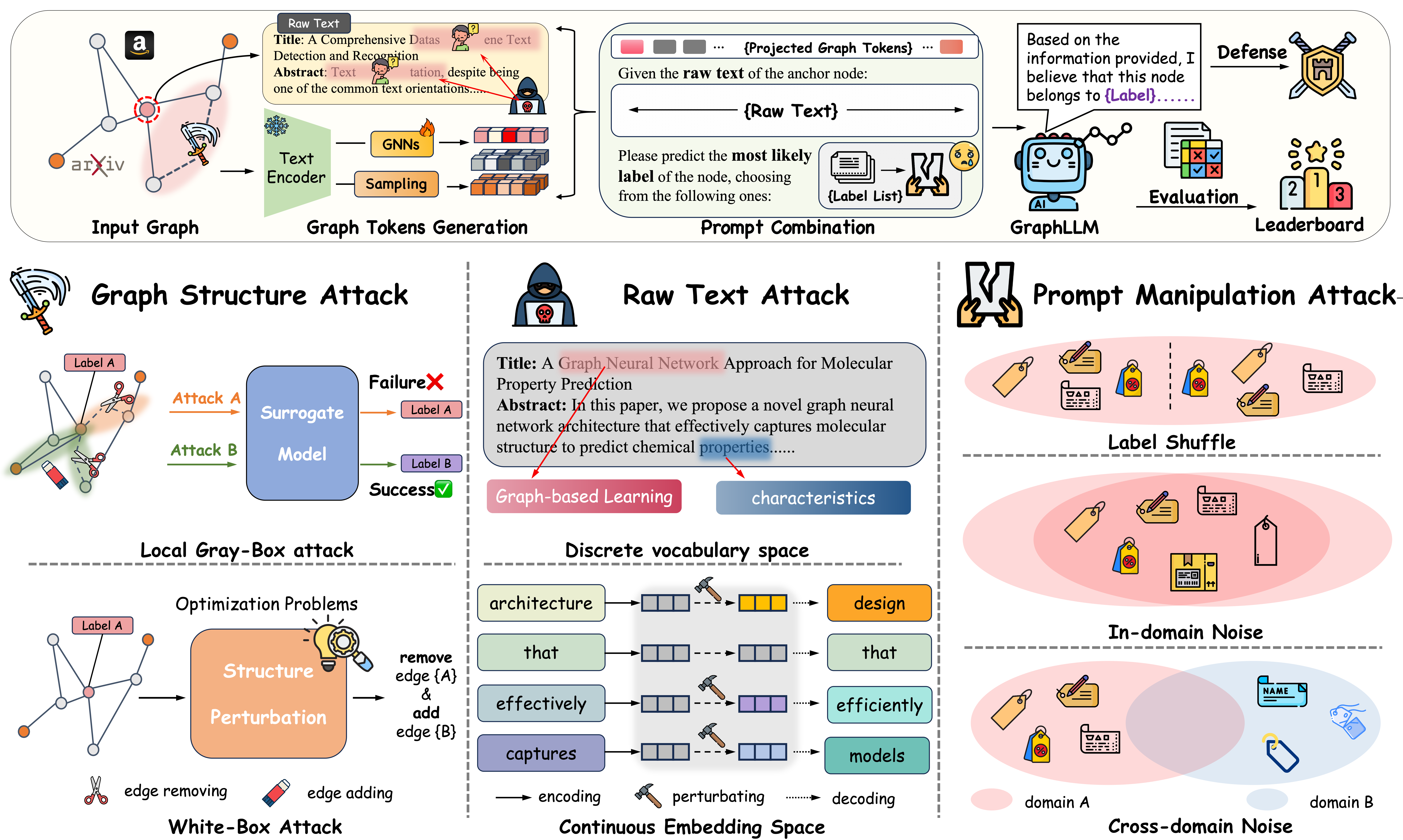}
    \caption{Overview of the TrustGLM benchmark. We evaluate the robustness of LLM-based graph learning models (a.k.a. GraphLLMs) across three distinct dimensions: text attacks, graph structure attacks, and prompt label attacks. For text and structure attacks, we implement state-of-the-art adversarial techniques to assess model performance. For prompt label attacks, we introduce a set of simple yet effective strategies to expose vulnerabilities in existing GraphLLMs. Additionally, for each attack type, we propose defense techniques that are empirically shown to enhance the robustness of GraphLLM methods.}
    \label{fig:general_architecture}
\end{figure*}

\section{Preliminary}
\textbf{Text-Attributed Graphs.} A Text-Attributed Graph (TAG) is defined as 
\(
G = (V, E, X, T),
\)
where \( V \) is the node set, \( E \subseteq V \times V \) is the edge set, and each node \( v_i \in V \) has an associated textual description \( t_i \in T \). The feature matrix \( X \in \mathbb{R}^{|V| \times d} \) is derived from \( T \), where \( x_i = \phi(t_i) \) is obtained via an encoding function \( \phi: T \to \mathbb{R}^d \). The adjacency matrix \( A \in \{0,1\}^{|V| \times |V|} \) encodes graph connectivity, with \( A_{ij} = 1 \) if \( (v_i, v_j) \in E \).

\noindent \textbf{Graph Large Language Model.} Graph Large Language Model extends standard LLMs by integrating structured node embeddings into the textual input. Given a target node \( v_i \) with textual description \( t_i \) and task-specific descriptions, its tokenized sequence is \( X_i = (x_i^1, x_i^2, \dots, x_i^m) \), where each \( x_i^j \) represents a token embedding. A graph encoder \( f_G \) computes node embeddings \( h_i = f_G(G, v_i) \), and selected neighbor embeddings \( H = \{ h_j \mid v_j \in \mathcal{N}(i) \} \) provide additional structural context. These embeddings are then concatenated to form an extended embedding sequence:
\(
Z_i = (h_{j_1}, h_{j_2}, \dots, h_{j_k}, x_i^1, x_i^2, \dots, x_i^m),
\)
where \( \{v_{j_1}, \dots, v_{j_k}\} \subseteq \mathcal{N}(i) \) are selected neighbors of \( v_i \). The final input consists of the standard textual tokens along with the graph-informed embedding sequence.
The LLM, parameterized by \( \theta \), generates an output sequence \( Y = (y_1, y_2, \dots, y_r) \), with the probability distribution modeled as  
\begin{equation}
p_{\theta}(Y \mid Z_i) = \prod_{k=1}^{r} p_{\theta}(y_k \mid Z_i, y_{1:k-1}).
\end{equation}
Here, \( p_{\theta}(y_k \mid Z_i, y_{1:k-1}) \) denotes the probability of generating \( y_k \) conditioned on prior tokens and the structured embedding sequence. By incorporating node and neighborhood embeddings directly into the model’s representation space, GraphLLMs effectively integrate relational information into LLM-based inference.


\section{T\lowercase{rust}GLM Benchmark}
In this section, we introduce TrustGLM, a benchmark designed to systematically evaluate the robustness of GraphLLMs against adversarial attacks. TrustGLM provides a standardized framework for assessing vulnerabilities in GraphLLMs across different attack vectors, facilitating the development of more resilient models. The overall framework of our benchmark is illustrated in Figure~\ref{fig:general_architecture}.
Furthermore, we propose a comprehensive taxonomy of GraphLLM attacks, categorizing them into Graph Structure Attacks (Section~\ref{sec:GSA}), Raw Text Attacks (Section~\ref{sec:RTA}), and Prompt Manipulation Attacks (Section~\ref{sec:PMA}). For each attack paradigm, we introduce corresponding defense strategies to mitigate their impact, ensuring a structured and systematic analysis of GraphLLM security.

\subsection{Graph Structure Attack-and-Defense}
\label{sec:GSA}
\subsubsection{Attack}
Graph structure attacks compromise GraphLLMs by strategically modifying the adjacency matrix 
\(A\), which encodes node connectivity. By adding or removing edges under a predefined budget, adversaries reshape neighborhoods in ways that mislead the model’s inference. In practice, such attacks can be \textit{local} (targeting specific nodes) or \textit{global} (aimed at degrading overall performance).

For our benchmark, we employ two representative methods: Nettack~\cite{zugner2018adversarial} and PRBCD~\cite{geisler2021prbcd}. Nettack is a local gray-box attack that iteratively modifies a target node’s surrounding subgraph to induce misclassification, guided by a surrogate model trained on accessible data. PRBCD is a more scalable white-box attack that can operate in both local and global settings by transforming discrete adjacency changes into a continuous optimization process; it updates edge blocks based on gradient information while ensuring the number of edge modifications stays within budget.

Because querying large language models repeatedly is computationally expensive, we generate structural perturbations \textit{before} inference using a surrogate model (e.g., GCN). This gray-box strategy ensures the perturbed graph effectively challenges GraphLLM robustness without excessive overhead.

\subsubsection{Defense.}
To counter structure attacks, we integrate adversarial training into the GraphLLM pipeline, focusing on two gradient-based defense approaches: \textit{FGSM}~\cite{goodfellow2014explaining} and \textit{PGD}~\cite{madry2018pgd}. Both methods generate adversarial perturbations, here, edge flips, or insertions, by computing gradients with respect to the adjacency matrix.
It is worth noting that graph structure learning is another way to defense structure attacks~\cite{zhou2023opengsl,fang2024gaugllm,guo2024learning}. However, they often suffer from scalability issue, so we do not include them.

\textit{FGSM} applies a single-step gradient update to produce fast adversarial samples, which are then combined with clean data during training. This process encourages the model to learn representations that remain stable under small but intentionally crafted structural changes. In contrast, \textit{PGD} iterates multiple times over the gradient and projection steps, gradually refining the perturbations to be both more potent and still within the modification budget. Training with these stronger samples makes the GraphLLM more robust to a wider range of structure-based attacks.


\subsection{Raw Text Attack-and-Defense}
\label{sec:RTA}
\subsubsection{Attack.}
Raw Text based adversarial attacks aim to create minimally perturbed textual inputs that retain the original semantic meaning yet cause misclassifications. Formally, given an original text sample \( x \) with label \( y \) and a classifier \( f \) such that \( f(x) = y\), we seek an adversarial \( x^* \) satisfying:
\begin{equation}
x^* = \underset{x'}{\arg\min}\, \Bigl\{ -G(x, x') \Bigr\} 
\quad
\text{s.t.}
\quad
f(x') \neq y,
\end{equation}
where \( G(\cdot,\cdot) \) measures semantic similarity. For GraphLLMs, we adopt a black-box paradigm, which consistently tests each model’s vulnerability without leveraging its internal parameters.

In our benchmark, we employ two hard-label black-box strategies: \textit{HLBB}~\cite{HLBB} and \textit{TextHoaxer}~\cite{Ye_Miao_Wang_Ma_2022}. Both iteratively refine adversarial examples via constrained optimization to maintain high similarity to the original text. \textit{HLBB} operates with decision-based feedback using only top predicted labels, whereas \textit{TextHoaxer} formulates perturbations in the continuous embedding space to achieve stronger and more efficient attacks. Notably, gradient-based white-box methods \cite{guo-etal-2021-gradient, guo2024learninggraphslargelanguage} presumes full access to model parameters, which falls outside the hard-label black-box setting we investigate and is therefore omitted from this study.

\subsubsection{Defense.}
To enhance model resilience, we integrate adversarial data augmentation into the training of GraphLLMs. Specifically, we first generate a perturbed version of each node’s text in the training set:
\begin{equation}
\mathcal{D}' = \{ (x_i^*, y_i) \,\mid\, (x_i, y_i) \in \mathcal{D} \}, 
\end{equation}
where \( x_i^* \) denotes an adversarially perturbed sample. We then create an augmented dataset 
\(
\mathcal{D}^{\ast} = \mathcal{D} \cup \mathcal{D}',
\)
using both clean and adversarial examples for training. By learning from these diverse textual inputs, GraphLLMs develop more robust representations, reducing their susceptibility to future text-based manipulations.

\begin{table*}[ht]
  \centering
  \caption{Raw Text Attack Results: Accuracy (Acc) and Attack Success Rate (ASR).}
  \label{tab:text_attack}
  \begin{tabular}{l l | c c | c c | c c}
    \toprule
    \multirow{2}{*}{\textbf{Dataset}} & \multirow{2}{*}{\textbf{Attack Method}} & 
    \multicolumn{2}{c|}{\textbf{GraphPrompter}} & 
    \multicolumn{2}{c|}{\textbf{LLaGA}} & 
    \multicolumn{2}{c}{\textbf{GraphTranslator}} \\
    \cmidrule(lr){3-4} \cmidrule(lr){5-6} \cmidrule(lr){7-8}
     & & \textbf{Acc} & \textbf{ASR} & \textbf{Acc} & \textbf{ASR} & \textbf{Acc} & \textbf{ASR} \\
    \midrule

    \multirow{3}{*}{\textbf{Cora}}
    & Original       
       & 67.71\%  & -  
       & 87.82\%  & - 
       & 24.54\%  & - \\
     & HLBB       
       & 45.90\% (\textminus32.2\%) & 32.20\%  
       & 57.30\% (\textminus34.8\%) & 34.75\%  
       & 18.60\% (\textminus24.2\%) & 24.19\% \\
     & TextHoaxer  
       & 41.51\% (\textminus38.7\%) & 38.70\%  
       & 46.74\% (\textminus46.8\%) & 46.77\%  
       & 20.58\% (\textminus16.1\%) & 16.12\% \\
    \midrule

    \multirow{3}{*}{\textbf{Pubmed}}
    & Original       
       & 94.35\%  & -  
       & 89.02\%  & - 
       & 64.10\%  & - \\
     & HLBB       
       & 77.65\% (\textminus17.7\%) & 17.70\%  
       & 66.23\% (\textminus25.6\%) & 25.60\%  
       & 56.95\% (\textminus11.2\%) & 11.15\% \\
     & TextHoaxer  
       & 72.17\% (\textminus23.5\%) & 23.50\%  
       & 62.83\% (\textminus29.4\%) & 29.42\%  
       & 50.66\% (\textminus21.0\%) & 20.96\% \\
    \midrule

    \multirow{3}{*}{\textbf{OGB-Products}}
    & Original       
       & 79.98\%  & -  
       & 73.08\%  & - 
       & 24.56\%  & - \\
     & HLBB       
       & 56.18\% (\textminus29.8\%) & 29.75\%  
       & 52.01\% (\textminus28.8\%) & 28.83\%        
       & 22.16\% (\textminus9.8\%)  & 9.76\% \\
     & TextHoaxer  
       & 59.59\% (\textminus25.5\%) & 25.49\%  
       & 45.71\% (\textminus37.4\%) & 37.44\%  
       & 20.82\% (\textminus15.2\%) & 15.21\% \\
    \midrule

    \multirow{3}{*}{\textbf{OGB-Arxiv}}
    & Original       
       & 72.68\%  & -  
       & 73.20\%  & - 
       & 15.40\%  & - \\
     & HLBB       
       & 62.53\% (\textminus14.0\%) & 13.96\%  
       & 61.50\% (\textminus16.0\%) & 15.98\%       
       & 14.20\% (\textminus7.8\%)  & 7.82\% \\
     & TextHoaxer  
       & 49.88\% (\textminus31.4\%) & 31.37\%  
       & 53.23\% (\textminus27.3\%) & 27.27\%  
       & 14.09\% (\textminus8.5\%) & 8.46\% \\
    \bottomrule
  \end{tabular}
\end{table*}


\begin{figure}[ht]
    \centering
    \includegraphics[width=\linewidth]{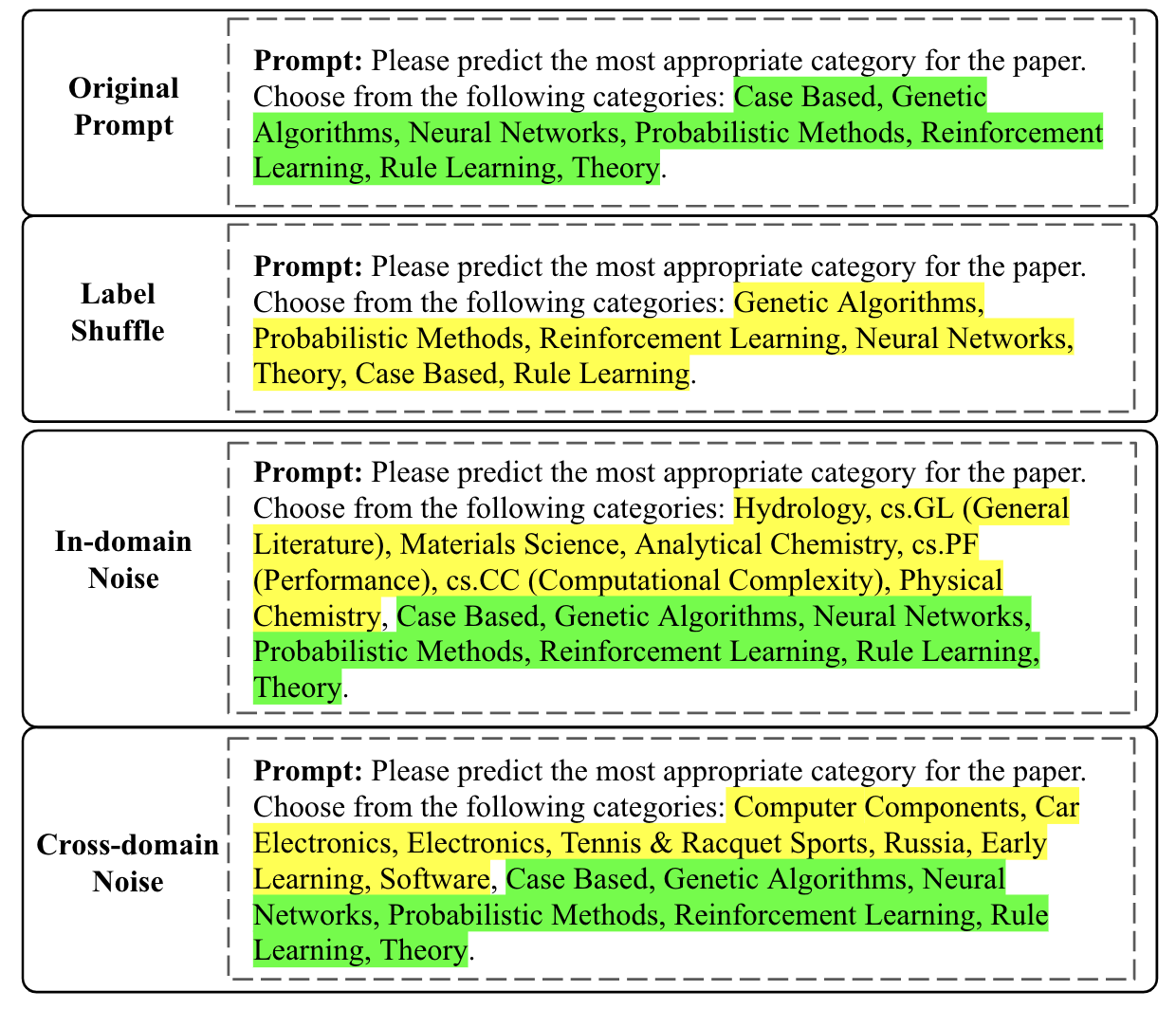}
    \caption{Three different kinds of Prompt Attacks.}
    \label{fig:promptattack}
\end{figure}

\subsection{Prompt Manipulation Attack-and-Defense}
\label{sec:PMA}
\subsubsection{Attack.}
When integrating LLMs for graph reasoning tasks such as classification, it is often necessary to provide relevant category information to help LLMs clearly identify classification objectives. Therefore, in this paper, Prompt Manipulation Attacks primarily target the provided label information.
Prompt manipulation attacks operate by modifying the structured input—label sets—so that the model’s predicted label becomes incorrect, even though the underlying node features or graph structure remain unchanged. Given a prompt \(P\) composed of an instruction \(I\) and a label set \(L = (l_1, l_2, \dots, l_m)\), an attacker applies a transformation $T$ to produce a new prompt $P' = T(P)$. This altered prompt retains a valid structure but disrupts how the model associates labels with the correct answer. The objective of the attacker is to ensure that the model's prediction under $P'$ deviates from the expected outcome:
\begin{equation}
\hat{l} = f(P') \quad \text{such that} \quad \hat{l} \neq l^*,
\end{equation}
where $l*$ is the correct label under original prompt $P$.

In our benchmark, we propose three principal manipulation types of prompt label attack: label shuffle attack, in-domain noise attack, cross-domain noise attack. Examples are shown in figure \ref{fig:promptattack}.

\paragraph{\textbf{Label Shuffle Attack.}}
A permutation function \(\pi: \{1,2,\ldots,m\} \to \{1,2,\ldots,m\}\) is applied to reorder labels:
\begin{equation}
    L' = \pi(L) \;=\; \bigl(l_{\pi(1)},\, l_{\pi(2)},\, \dots,\, l_{\pi(m)}\bigr).
\end{equation}
Such reordering disrupts any positional or sequential cues that the model may rely on for prediction.

\paragraph{\textbf{In-Domain Noise Attack.}}
Additional labels from a \emph{related} set \(N_{\text{in}}\) are injected into \(L\):
\begin{equation}
    L' = L \cup N_{\text{in}}, 
    \quad 
    N_{\text{in}} \subseteq \mathcal{L}_{\text{origin-domain}},
\end{equation}
where \(\mathcal{L}_{\text{domain}}\) represents labels relevant to the original task domain. This expands the label set in a manner that still appears contextually plausible, increasing confusion during prediction.

\paragraph{\textbf{Cross-Domain Noise Attack.}}
Labels from an \emph{unrelated} domain \(N_{\text{cross}}\) are added:
\begin{equation}
    L' = L \cup N_{\text{cross}},
\end{equation}
where
\begin{equation}
    N_{\text{cross}} \subseteq \mathcal{L}_{\text{other-domain}}, 
    \quad
    \mathcal{L}_{\text{other-domain}} \cap \mathcal{L}_{\text{origin-domain}} = \emptyset.
\end{equation}
By introducing completely foreign or semantically unrelated labels, the model’s decision boundaries are further challenged, often causing significant misclassifications.

\subsubsection{Defense.} 
To improve the robustness of GraphLLMs against Prompt Manipulation Attacks, we introduce adversarial training techniques tailored to different types of prompt perturbations. For label shuffle attacks, we augment the training process by randomly permuting the label order within prompts, ensuring that the model learns to be invariant to changes in label positions. For in-domain and cross-domain noise attacks, we incorporate additional related labels into training prompts to expose the model to broader category variations, thereby reducing sensitivity to label set expansion.

\section{Experiments}

In this section, we conduct extensive experiments under our benchmark setting, aiming to address the following research questions:
\begin{itemize}
    \item \textbf{RQ1}: How severely can text, structure, and label-based adversarial manipulations degrade GraphLLMs?
    \item \textbf{RQ2}: Which inherent design or domain factors fundamentally shape GraphLLMs’ susceptibility to diverse threats?
    \item \textbf{RQ3}: Are data augmentation and adversarial training sufficient to preserve accuracy when facing adversarial inputs?
    \item \textbf{RQ4}: How do classical adversarial training methods compare with one another in enhancing the performance of GraphLLMs under graph structure attacks?
    \item \textbf{RQ5}: How do the quantity and position of label noise affect the effectiveness of the attack?

\end{itemize}
\subsection{Experiment Configuration}
\subsubsection{\textbf{Datasets}}
Our experiments are conducted on two types of datasets: Citation Networks and Amazon Review. For Citation Networks, we employed Cora~\cite{mccallum2000automating}, PubMed ~\cite{sen2008collective}, 
2000), and OGB-Arxiv~\cite{hu2020open}. Amazon Review datasets include the subset of OGB-Products~\cite{hu2020open}, Amazon-Computers and Amazon-Sports~\cite{shchur2018amazon}. While Prompt Label attacks are applied to all datasets, Language and Structure attacks target Citation Networks and OGB-products, representing the Amazon Review Graph. The statistics of all the datasets can be found in Table~\ref{tab:freq}.
\begin{table}
  \caption{Dataset Statistics.}
  \label{tab:freq}
  \begin{tabular}{cccc}
    \toprule
    Dataset & \#Nodes & \#Edges & \#Classes \\
    \midrule
    Cora & 2708 & 10,556 & 7 \\
    Pubmed & 19,717 & 88,648 & 3 \\
    OGB-Arxiv & 169,343 & 2,315,598 & 40 \\
    OGB-Products (subset) & 54,025 & 144,638 & 47 \\
    Amazon-Computers & 87,229 & 721,107 & 10 \\
    Amazon-Sports & 173,055 & 1,946,555 & 13 \\
    \bottomrule
  \end{tabular}
\end{table}


\subsubsection{\textbf{Victim GraphLLM}}
We employ LLaGA \cite{chen2024llaga}, GraphPrompter \cite{liu2024can}, and GraphTranslator \cite{zhang2024graphtranslator} as our victim GLLMs.

\begin{table*}[ht]
  \centering
  \caption{Graph Structure Attack Results: Accuracy (Acc) and Attack Success Rate (ASR).}
  \label{tab:strcattack}
  \begin{tabular}{l l | c c | c c | c c}
    \toprule
    \multirow{2}{*}{\textbf{Dataset}} & \multirow{2}{*}{\textbf{Method}} & 
    \multicolumn{2}{c|}{\textbf{LLaGA}} & 
    \multicolumn{2}{c|}{\textbf{GraphPrompter}} &
    \multicolumn{2}{c}{\textbf{GraphTranslator}} \\
    \cmidrule(lr){3-4} \cmidrule(lr){5-6} \cmidrule(lr){7-8}
     & & \textbf{Acc} & \textbf{ASR} & \textbf{Acc} & \textbf{ASR} & \textbf{Acc} & \textbf{ASR} \\
    \midrule

    \multirow{4}{*}{\textbf{Cora}} 
     & Original       & 87.82\% & -            & 67.71\% & -            & 24.54\% & - \\
     & Nettack        & 50.00\% (\textminus43.07\%) & 41.33\% & 60.52\% (\textminus10.62\%) & 14.02\% & 24.91\% (+1.51\%) & 6.28\% \\
     & PRBCD (local)  & 62.18\% (\textminus29.20\%) & 27.68\% & 60.52\% (\textminus10.62\%) & 15.50\% & 19.56\% (\textminus20.31\%) & 39.56\% \\
     & PRBCD (global) & 74.91\% (\textminus14.69\%) & 14.39\% & 64.94\% (\textminus4.09\%)  & 9.04\%  & 19.00\% (\textminus22.59\%) & 39.19\% \\
    \midrule

    \multirow{4}{*}{\textbf{Pubmed}} 
     & Original       & 89.02\% & -     & 94.35\% & -     & 64.10\% & - \\
     & Nettack        & 56.57\% (\textminus36.45\%) & 33.67\% & 94.14\% (\textminus0.22\%)  & 0.51\%  & 64.07\% (\textminus0.05\%)  & 0.48\% \\
     & PRBCD (local)  & 78.35\% (\textminus11.98\%) & 12.55\% & 94.19\% (\textminus0.17\%)  & 0.46\%  & 57.96\% (\textminus9.58\%)  & 15.32\% \\
     & PRBCD (global) & 81.85\% (\textminus8.06\%)  & 9.76\%  & 94.30\% (\textminus0.05\%)  & 0.33\%  & 57.91\% (\textminus9.65\%)  & 15.29\% \\
    \midrule

    \multirow{4}{*}{\textbf{OGB-Products}} 
     & Original       & 73.08\% & -     & 79.98\% & -     & 24.56\% & - \\
     & Nettack        & 30.16\% (\textminus58.76\%) & 59.16\% & 77.78\% (\textminus2.75\%)  & 5.46\%  & 18.62\% (\textminus24.19\%) & 35.33\% \\
     & PRBCD (local)  & 62.44\% (\textminus14.56\%) & 15.92\% & 78.66\% (\textminus1.65\%)  & 4.68\%  & 18.68\% (\textminus23.94\%) & 35.41\% \\
     & PRBCD (global) & 63.46\% (\textminus13.16\%) & 15.32\% & 79.20\% (\textminus0.98\%)  & 3.38\%  & 18.64\% (\textminus24.11\%) & 35.45\% \\
    \midrule

    \multirow{3}{*}{\textbf{OGB-Arxiv}} 
     & Original       & 73.20\% & -     & 72.68\% & -     & 15.40\% & - \\
     & PRBCD (local)  & 66.26\% (\textminus9.49\%)  & 14.44\% & 69.58\% (\textminus4.27\%)  & 11.84\% & 15.70\% (+1.95\%) & 3.04\% \\
     & PRBCD (global) & 45.72\% (\textminus37.55\%) & 33.66\% & 70.84\% (\textminus2.53\%)  & 13.28\% & 15.48\% (+0.51\%) & 4.32\% \\
    \bottomrule
  \end{tabular}
\end{table*}

\subsubsection{\textbf{Evaluation Metric}}
To assess the effectiveness of our attacks, we report the \textbf{Attack Success Rate (ASR)}, defined as the proportion of originally correctly classified nodes whose predictions change after perturbation. For graph attacks, ASR is computed differently. instead of counting only successfully misclassified nodes, the numerator includes all targeted nodes, as some attack methods operate globally rather than on an individual node basis. Additionally, we report the \textbf{post-attack classification accuracy} to provide insight into the overall degradation in model performance under adversarial perturbations.

\subsection{Overall Attack Evaluation (RQ1)}
\subsubsection{\textbf{Raw Text Attack}}

To assess the robustness of GraphLLMs against textual perturbations, we conducted text attacks on the victim models. We used HLBB and TextHoaxer as attack methods, details can be found in Appendix A. To ensure a fair and efficient evaluation, we restricted our attacks to nodes that were originally correctly classified. Given computational constraints, we sampled 10\% of these correctly classified test nodes and repeated the attack process three times to obtain stable results. The attack outcomes are summarized in Table \ref{tab:text_attack}. Below, we highlight two key observations:

~\ding{192}~\textbf{GraphLLMs are vulnerable to adversarial text attacks.} Across all datasets, accuracy reductions are observed after both HLBB and TextHoaxer attacks. For instance, in the Cora dataset, GraphPrompter’s accuracy drops from 67.71\% to 41.51\% and LLaGA’s accuracy drops from 87.82\% to 46.74\% , demonstrating that these attacks can effectively degrade model performance.

~\ding{193} \textbf{Models that rely more on textual features for classification tend to be more vulnerable, whereas models with stronger graph priors exhibit greater resilience to text attacks.} LLaGA, which processes node sequences from text, experiences the highest ASR values, reaching 46.77\% on Cora and 37.44\% on Product under TextHoaxer. In contrast, GraphTranslator, which incorporates stronger graph-based reasoning, maintains lower ASR values, with 24.19\% on Cora and 15.21\% on Product under TextHoaxer, suggesting that integrating richer graph structures enhances robustness against textual perturbations.

\subsubsection{\textbf{Graph Structure Attack}}
To test GraphLLM's robustness against graph structure perturbations, we applied Netattack and PRBCD graph attack to victim GraphLLMs. 
We first trained a surrogate GCN model using the same training data as the GraphLLMs. Next, we applied both Nettack and PRBCD to the surrogate model to generate perturbed adjacency matrices. In the case of Nettack and local PRBCD, attacks were conducted on individual test nodes during inference, producing a distinct perturbed adjacency matrix for each node to manipulate its predicted label. Conversely, global PRBCD modified the graph structure collectively, aiming to degrade overall model accuracy by generating a single perturbed adjacency matrix for all test nodes. These modified adjacency matrices were then used for inference with the GraphLLMs. Specifically, for LLaGA, node sequences were sampled from the perturbed structures before inference. For GraphPrompter, the K-hop subgraph of each test node was extracted from the perturbed structures before passing through the GNN encoder. For GraphTranslator, test nodes were encoded using GraphSAGE based on the perturbed graph structure. Results of graph attack are shown in table \ref{tab:strcattack}. We made the following observations:

~\ding{194} \textbf{GraphLLMs exhibit varying degrees of vulnerability to structure attacks, with some models experiencing substantial performance degradation.} LLaGA is the most affected, showing the largest accuracy drops across datasets, such as a 43.07\% decrease on Cora under Nettack and a 37.55\% drop on OGB-Arxiv under PRBCD (global). In contrast, GraphPrompter and GraphTranslator demonstrate greater resilience, particularly on Pubmed, where accuracy reductions remain minimal (e.g., 0.22\% for GraphPrompter under Nettack). This suggests that GraphLLMs with stronger reliance on structural information suffer more under structure-based perturbations.

\begin{table*}[ht]
  \centering
  \caption{Prompt Manipulation Attack Results: Accuracy (Acc). ``-'' indicates that no specific label candidates were provided in the original dataset, thereby rendering the prompt manipulation attack inapplicable.}
  \label{tab:prompt_label_attack}
  \begin{tabular}{l l c c c c }
    \toprule
    \textbf{Model} & \textbf{Dataset} & 
    \textbf{Original} & \textbf{Shuffle} & 
    \makecell{\textbf{In-Domain Noise}} & 
    \makecell{\textbf{Cross-Domain Noise}} \\
    \midrule

    \multirow{6}{*}{\textbf{GraphPrompter}} 
     & Amazon-Sports    & 91.99\% & 78.86\% (\textminus14.27\%) & 80.41\% (\textminus12.59\%) & 66.86\% (\textminus27.33\%) \\
     & Amazon-Computers & 78.31\% & 57.78\% (\textminus26.23\%) & 67.30\% (\textminus14.05\%) & 20.89\% (\textminus73.35\%) \\
     & OGB-Products     & 79.98\% & 63.66\% (\textminus20.40\%) & 45.92\% (\textminus42.58\%) & 34.20\% (\textminus57.23\%) \\
     & Cora             & 67.71\% & 53.14\% (\textminus21.51\%) & 52.40\% (\textminus22.60\%) & 60.33\% (\textminus10.89\%) \\
     & Pubmed           & 94.35\% & 94.57\% (+0.23\%)            & 93.05\% (\textminus1.38\%)  & 93.38\% (\textminus1.03\%)  \\
     & OGB-Arxiv        & 72.68\% & -                            & -                            & -                            \\
    \midrule

    \multirow{6}{*}{\textbf{LLaGA}} 
     & Amazon-Sports    & 91.32\% & 83.08\% (\textminus9.02\%)   & 84.48\% (\textminus7.49\%)  & 82.16\% (\textminus10.03\%) \\
     & Amazon-Computers & 87.41\% & 81.85\% (\textminus6.36\%)   & 82.64\% (\textminus5.45\%)  & 77.70\% (\textminus11.11\%) \\
     & OGB-Products     & 73.08\% & 73.42\% (+0.46\%)            & 70.28\% (\textminus3.83\%)  & 67.30\% (\textminus7.91\%)  \\
     & Cora             & 87.82\% & 87.45\% (\textminus0.42\%)   & 85.42\% (\textminus2.73\%)  & 87.27\% (\textminus0.63\%)  \\
     & Pubmed           & 89.02\% & 87.65\% (\textminus1.54\%)   & 85.37\% (\textminus4.10\%)  & 85.52\% (\textminus3.93\%)  \\
     & OGB-Arxiv        & 73.20\% & 69.58\% (\textminus4.95\%)   & 73.12\% (\textminus0.11\%)  & 72.62\% (\textminus0.79\%)  \\
    \bottomrule
  \end{tabular}
  \vspace{-8pt}
\end{table*}

~\ding{195} \textbf{Local attacks (Nettack, PRBCD local) tend to be more damaging on structure-sensitive models, while global PRBCD attacks show more effectiveness at degrading overall model accuracy. } For example, on the Cora dataset using LLaGA, Nettack reduces accuracy by 43.07\% with an ASR of 41.33\%, whereas global PRBCD decreases accuracy by 14.69\% with only a 14.39\% ASR. This indicates that, in some cases, targeted (local) attacks can be far more devastating. Meanwhile, GraphPrompter remains relatively stable under PRBCD attacks, with accuracy drops of only 0.98\% on OGB-Products under PRBCD (global), highlighting its robustness against structure perturbations.

\subsubsection{\textbf{Prompt Manipulation Attack}}
We evaluated model's robustness against prompt purbations in this section. 
In the Label Shuffle Attack, the order of label candidates in the prompt is randomized using a fixed seed. For the Label In-Domain Noise Attack, additional labels from related domains are shuffled, selected with a fixed seed, and added to the candidate list in the prompt. Specifically, for the Amazon Review datasets, labels from Amazon-Children, Amazon-Photo, Amazon-History, Amazon-Sports, and Amazon-Computers are incorporated; for the Citation Networks, labels from Cora, Pubmed, OGB-Arxiv, and 40 randomly selected MAG labels are used. In the Cross-Domain Noise Attack, labels from a different domain are added similarly—using labels from Cora, Pubmed, and OGB-Arxiv as noise for the Amazon Review dataset, and Amazon dataset labels for the Citation Networks.

We evaluate both noise quantity and position. For quantity, we experiment with extra labels equal to 50\% and 100\% of the original candidate set; for position, noise is added either before or after the original candidates. The reported results correspond to 100\% noise added at the front, with additional results provided in the ablation study. Results of prompt label attack are shown in table \ref{tab:prompt_label_attack}, from which we make the following observations:
\begin{figure}[ht]
  \centering
  \includegraphics[width=0.5\textwidth]{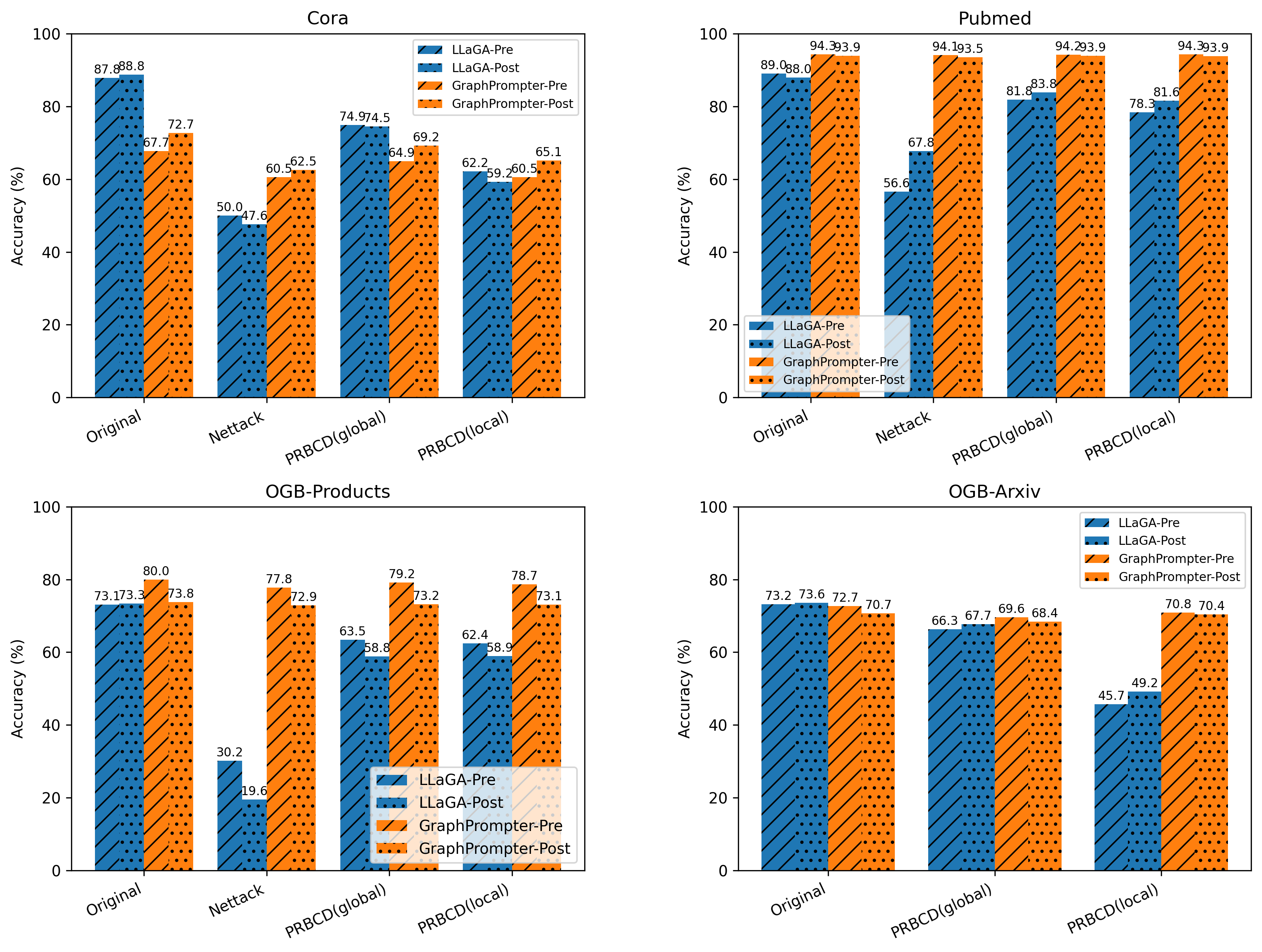}
  \caption{Comparison of model performance before and after FGSM adversarial training.}
  \label{fig:model_pre_post_hatch-1}
  \vspace{-8pt}
\end{figure}

\begin{table*}[ht]
  \centering
  \caption{Label Noise Training Results on \textbf{Cora} (Accuracy).}
  \label{tab:label_noise_training}
  \renewcommand{\arraystretch}{1.2}
  \resizebox{\textwidth}{!}{  
  \begin{tabular}{l ccc cc cc}
    \toprule
    \textbf{Model} 
    & \multicolumn{3}{c}{\textbf{No Defense}} 
    & \multicolumn{2}{c}{\textbf{In-Domain Defense}} 
    & \multicolumn{2}{c}{\textbf{Cross-Domain Defense}} \\
    \cmidrule(lr){2-4} \cmidrule(lr){5-6} \cmidrule(lr){7-8}
    & \textbf{Orig.} & \textbf{In-Domain Noise} & \textbf{Cross-Domain Noise}
    & \textbf{Orig.} & \textbf{In-Domain Noise}
    & \textbf{Orig.} & \textbf{Cross-Domain Noise} \\
    \midrule
    \textbf{LLaGA}         
    & 87.82\% & 85.42\% & 87.27\% 
    & 88.01\% & 89.11\% 
    & 88.93\% & 89.11\% \\
    \textbf{GraphPrompter} 
    & 67.71\% & 52.40\% & 60.33\% 
    & 66.24\% & 65.13\% 
    & 69.19\% & 69.37\% \\
    \bottomrule
  \end{tabular}
  }
  \vspace{-8pt}
\end{table*}

~\ding{196}~\textbf{GraphLLMs are generally vulnerable to prompt label perturbations.} GraphPrompter exhibits significant performance drops under label shuffling and in-domain noise attacks, with accuracy decreasing by up to 27.42\% (Amazon Computers). In contrast, LLaGA remains highly robust, with only marginal accuracy reductions in most cases. Notably, on Cora, Pubmed, and Arxiv, LLaGA's accuracy remains above 85\% even under the strongest cross-domain noise attack, indicating its resilience to label perturbations.

~\ding{197}~\textbf{Cross-domain noise is the most disruptive prompt label attack, particularly for models that depend more on textual label associations.} GraphPrompter's performance is severely degraded, especially on Amazon datasets, where accuracy drops by 57.42\% (Amazon Computers), indicating high sensitivity to label space manipulation. Meanwhile, LLaGA exhibits significantly lower performance degradation, highlighting its robustness to adversarial prompt modifications and its stronger reliance on graph priors.

\subsection{Comparative Evaluation of Different Attack Approaches (RQ2)}
\begin{figure*}[ht]
  \centering
  \begin{subfigure}[t]{0.49\textwidth}
    \centering
    \includegraphics[width=\linewidth]{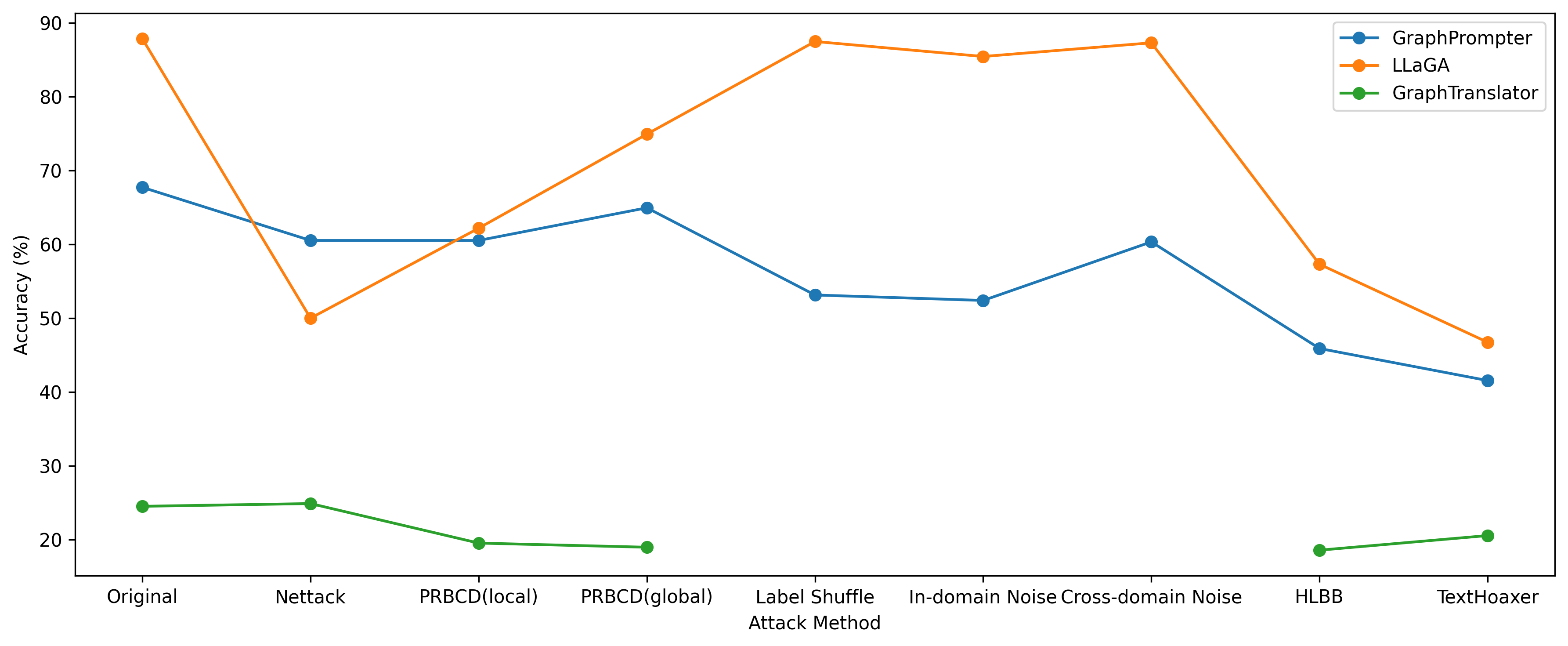}
    \caption{Cora}
    \label{fig:general_comparison_cora}
  \end{subfigure}
  \hfill
  \begin{subfigure}[t]{0.49\textwidth}
    \centering
    \includegraphics[width=\linewidth]{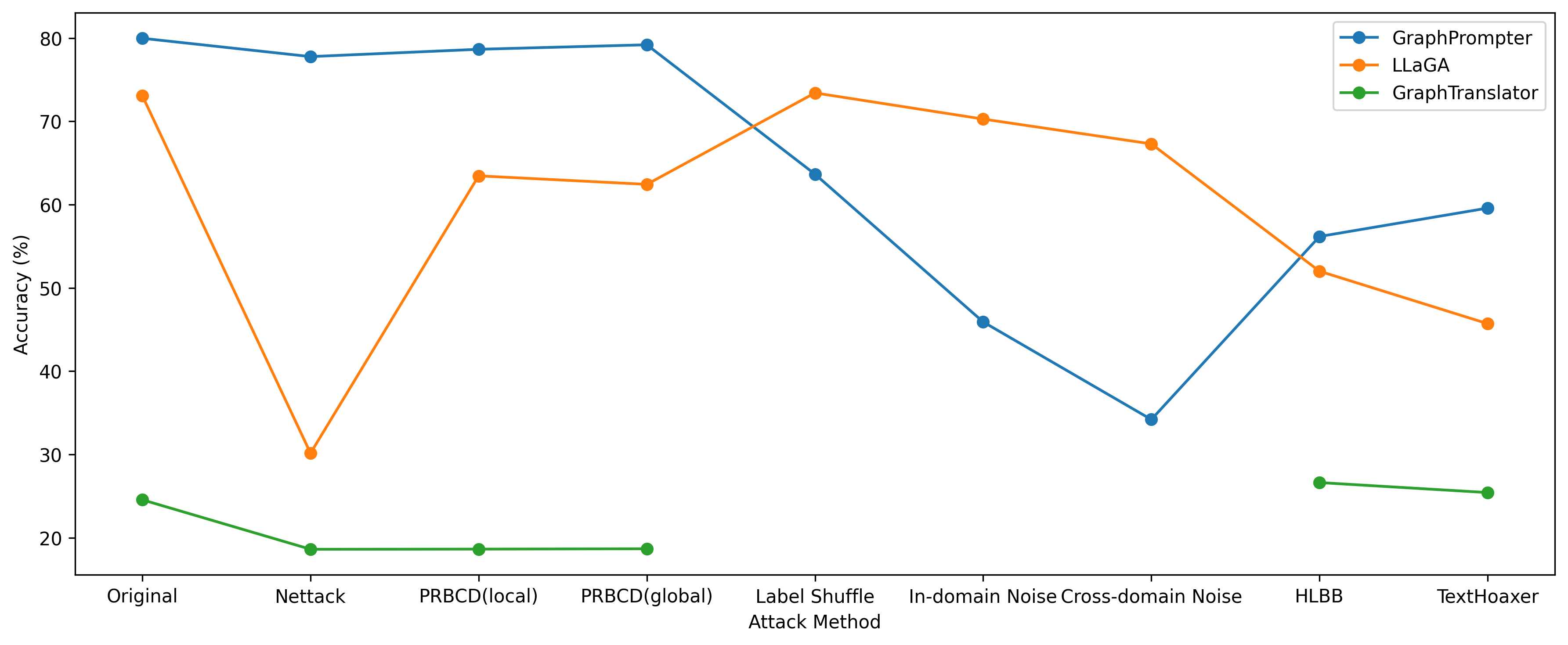}
    \caption{OGB-Products}
    \label{fig:general_comparison_products}
  \end{subfigure}
  \caption{Comparative evaluation of different attack approaches on Cora and OGB-Products.}
  \label{fig:general_comparison}
\end{figure*}

We compared various attack methods under the same datasets (Cora, Products) and setups. As shown in Figures \ref{fig:general_comparison_cora}\&\ref {fig:general_comparison_products}, we conclude that 
~\ding{198} \textbf{Text-based attacks are the most effective across models, causing the most significant accuracy drops, while structural and prompt label attacks exhibit varied impact depending on the model’s reliance on graph versus textual features.} Models that heavily depend on textual input, such as GraphPrompter, suffer substantial performance degradation under text attacks like HLBB and TextHoaxer, whereas models with stronger graph priors, like LLaGA, demonstrate greater resilience to such attacks but remain susceptible to structural perturbations like Nettack and PRBCD. Meanwhile, prompt label attacks have a more moderate yet consistent impact across models, indicating that while label perturbations introduce noise, they do not fundamentally disrupt the learned representations as severely as text or structural modifications. These findings highlight the necessity for model-specific defense strategies, particularly reinforcing robustness against text-based adversarial manipulations in GraphLLMs.



\subsection{How to Defense GraphLLMs Against Text, Structural, and Prompt Attacks?(RQ3)}
Beyond the attacks, we also investigate several strategies to defense existing GraphLLMs. Specifically, we used data augmentation and adversarial training to train GraphLLMs. Figure \ref{fig:model_pre_post_hatch-1} and Table \ref{tab:cora_texthoaxer} in Appendix summarize the results, from which we made the following observations for different attack perspectives.  We conclude that ~\ding{199} \textbf{While defense strategies improve model robustness, their effectiveness varies across different attack types and models, indicating that no single defense method is sufficient to fully mitigate all adversarial threats.}
\subsubsection{\textbf{Raw Text Attack Defense}}
We performed data-augmentation training with TextHoaxer on GraphPrompter and LLaGA. We attacked the training set and used the successfully attacked data as data augmentation during the training period. Then we tested the training using the same setting as in section 4.1.1. The results are reported in Table \ref{tab:cora_texthoaxer}, from which we observe that \textit{data augmentation training improves GraphPrompter’s robustness to text attacks but does not fully restore its original performance}. This suggests that data augmentation alone is insufficient to completely mitigate adversarial effects, highlighting the urgent need for more advanced defense strategies—such as specialized adversarial training objectives or enhanced regularization techniques tailored for GraphLLMs.
\begin{table}[ht]
  \centering
  \caption{Model performance before and after data augmentation training on Cora (accuracy).}
  \label{tab:cora_texthoaxer}
  \begin{tabular}{l | c c c}
    \toprule
    \textbf{Model} & \textbf{Orig.} & \textbf{Before-Train} & \textbf{Post-Train} \\
    \midrule
    GraphPrompter & 67.71\% & 41.51\% & 49.82\% \\
    LLaGA & 87.82\% & 46.74\% & 67.88\% \\
    \bottomrule
  \end{tabular}
\end{table}

\subsubsection{\textbf{Graph Structure Attack Defense}}
Adversarial training is applied to both LLaGA and GraphPrompter. For LLaGA, only the projector is fine-tuned during training with all other components kept frozen, while for GraphPrompter, the GNN and projector are jointly optimized, thereby enhancing the efficiency of the defense training process. Specifically, to compute the adversarial losses, for LLaGA, we generate adversarial examples by adding perturbations to the graph embedding input of the projector, whereas for GraphPrompter, adversarial perturbations are applied to the GNN input.
The results here are obtained by FGSM (see Figure~\ref{fig:model_pre_post_hatch-1}), while the results of PGD are included in the ablation study.

The impact of adversarial training varies across models and datasets. It improves performance for LLaGA on Pubmed and OGB-Arxiv, as well as for GraphPrompter on Cora. Other cases show only minor performance degradation—except for OGB-Products, where both models suffer significant declines. Overall, adversarial training leads to model- and dataset-specific trade-offs, underscoring the need for stronger defense strategies.

\subsubsection{\textbf{Prompt Manipulation Attack Defense}}
In Label shuffle training, we randomly shuffle the candidate labels in prompt for each sample to defense the targeted GraphLLMs. For label noise training, we randomly add noises to the candidate labels per sample. According to the results that are shown in Table \ref{tab:label_noise_training} and Table \ref{tab:label_shuffle_training}, we find that \ding{200} \textbf{Label defense methods significantly improve the performance of GraphLLMs against prompt manipulation attacks.} In our experiments, label shuffle training enhanced model performance on all datasets except for LLaGA on OGB-Products, where the shuffle attack was ineffective. Notably, label shuffle training increased the performance of GraphPrompter on Amazon-Computers under attack by 18.52\%. Additionally, label noise training improved the performance of both models on the Cora dataset under both in-domain and cross-domain noise attack. There results demonstrate our motivation to explore prompt defense strategies via random label shuffle training. 

\begin{table}[htbp]
    \centering
    \caption{Label Shuffle Training Results (accuracy). ``--'' indicates that no specific label candidates were provided in the prompt, thereby rendering the prompt manipulation attack and defense inapplicable.}
    \label{tab:label_shuffle_training}
    \renewcommand{\arraystretch}{1.2}
    \resizebox{\columnwidth}{!}{%
    \begin{tabular}{ll|cc|cc}
        \toprule
        \multirow{2}{*}{\textbf{Model}} & \multirow{2}{*}{\textbf{Dataset}} & \multicolumn{2}{c|}{\textbf{No Label Shuffle Training}} & \multicolumn{2}{c}{\textbf{Label Shuffle Training}} \\
        \cmidrule(lr){3-4} \cmidrule(lr){5-6}
        & & Original & Shuffle & Original & Shuffle \\
        \midrule
        
        \multirow{6}{*}{LLaGA} 
        & Cora              & 87.82\% & 87.45\% & 88.56\% & 89.48\% \\
        & Pubmed            & 89.02\% & 87.65\% & 89.81\% & 89.20\% \\
        & OGB-Products      & 73.08\% & 73.42\% & 73.42\% & 72.18\% \\
        & Amazon-Computers  & 87.41\% & 81.85\% & 87.36\% & 87.51\% \\
        & Amazon-Sports     & 91.32\% & 83.08\% & 91.04\% & 91.09\% \\
        & OGB-Arxiv         & 73.20\% & 69.58\% & 73.44\% & 73.60\% \\
        \midrule
        
        \multirow{6}{*}{GraphPrompter} 
        & Cora              & 67.71\% & 53.14\% & 69.19\% & 69.93\% \\
        & Pubmed            & 94.35\% & 94.57\% & 94.70\% & 94.62\% \\
        & OGB-Products      & 79.98\% & 63.66\% & 77.26\% & 77.46\% \\
        & Amazon-Computers  & 78.31\% & 57.78\% & 74.22\% & 76.30\% \\
        & Amazon-Sports     & 91.99\% & 78.86\% & 86.00\% & 87.18\% \\
        & OGB-Arxiv         & 72.68\% & --      & --      & -- \\
        \bottomrule
    \end{tabular}
    }
\end{table}

\subsection{Ablation Study (RQ4, RQ5)}
In this section, we conduct further experiments to analyze how key attack and defense configurations influence the performance of GraphLLMs.
\subsubsection{FGSM and PGD (\textbf{RQ4})}
After comparing the adversarial training results of FGSM and PGD with those presented in Table \ref{tab:defense_structure} of Appendix, we observe that their effectiveness varies across different attack methods and models, although their overall performance is generally comparable. Specifically, for the Cora dataset, PGD increases LLaGA’s accuracy under the PRBCD local attack to 62.55\%, whereas FGSM reduces it to 59.23\%. In contrast, for GraphPrompter under the PRBCD global attack, FGSM boosts performance to 69.19\%, outperforming PGD's improvement, which reaches only 68.27\%.

\begin{table}[ht]
\centering
\caption{Comparison of GraphLLMs trained with FGSM and PGD against graph structure attack on Cora (Accuracy).}
\label{tab:defense_structure}
\renewcommand{\arraystretch}{1.2}
\resizebox{\columnwidth}{!}{%
\begin{tabular}{c|ccc|ccc}
\toprule
\multirow{2}{*}{\textbf{Attack}} &
\multicolumn{3}{c|}{\textbf{LLaGA}} &
\multicolumn{3}{c}{\textbf{GraphPrompter}} \\
\cmidrule(lr){2-4} \cmidrule(lr){5-7}
& \textbf{Orig.} & \textbf{FGSM} & \textbf{PGD} 
& \textbf{Orig.} & \textbf{FGSM} & \textbf{PGD} \\
\midrule

Original
  & 87.82\% & 88.75\% (+1.06\%) & 88.19\% (+0.42\%) 
  & 67.71\% & 72.69\% (+7.35\%) & 70.66\% (+4.36\%) \\
Nettack
  & 50.00\% & 47.60\% (\textminus4.80\%) & 36.53\% (\textminus26.94\%) 
  & 60.52\% & 62.55\% (+3.36\%) & 66.05\% (+9.14\%) \\
PRBCD (global)
  & 74.91\% & 74.54\% (\textminus0.49\%) & 73.43\% (\textminus1.98\%) 
  & 64.94\% & 69.19\% (+6.54\%) & 68.27\% (+5.13\%) \\
PRBCD (local)
  & 62.18\% & 59.23\% (\textminus4.75\%) & 62.55\% (+0.60\%) 
  & 60.52\% & 65.13\% (+7.62\%) & 65.68\% (+0.84\%) \\
\bottomrule
\end{tabular}
}
\end{table}
\subsubsection{Label noise attack with different noise quantities and positions (\textbf{RQ5})}
From table \ref{tab:comparison} in Appendix, we see that label noise attack effectiveness is largely influenced by both the noise quantity and position; yet~\ding{201} \textbf{adding noise to the beginning of the candidate labels is more effective than appending it to the end, and a noise quantity of 100\% results in a stronger attack than 50\%.} For both models, across nearly all datasets, when the noise position is held constant, 100\% noise degrades model performance more than 50\% noise. For example, on the OGB-Products dataset with GraphPrompter, applying 100\% front cross-domain noise reduces performance by an additional 34.64\% compared to 50\% noise. Similarly, when the noise quantity is fixed, placing noise at the beginning generally causes a greater performance drop than appending it at the end; for instance, on the Amazon-Sports dataset, 50\% noise applied at the front decreases performance by 4.55\% more than when applied at the end.
\section{Conclusions}
This work introduces TrustGLM, a benchmark designed to evaluate the robustness of Graph Large Language Models (GraphLLMs) against adversarial attacks. We systematically examined vulnerabilities from three perspectives—text, graph structure, and prompt label attacks—and found that these threats significantly degrade performance across multiple datasets. To counter these vulnerabilities, we investigated defense techniques including adversarial data augmentation and adversarial training, which improved robustness yet did not completely eliminate the risks. Our findings highlight the urgent need for more effective defenses and lay a solid foundation for future research in securing GraphLLMs.

\bibliographystyle{ACM-Reference-Format}
\balance
\bibliography{trustglm-citation.bib}

\newpage
\appendix

\section{Attack and Defense methods}
\subsection{Text Attack}
\subsubsection{HLBB}
This method tackles adversarial attacks by employing a decision-based strategy that requires only the top predicted label from the target model. Their formulation also follows a constrained optimization approach where the objective is to maximize the semantic similarity $S(x, x^\star)$ between the original text and the adversarial example, while ensuring that the adversarial criterion $C$ (i.e., $f(x^\star) \neq f(x)$) holds. This can be expressed as:

\begin{equation}
x^* = \arg\max_{x'} \Bigl\{ S(x,x') \Bigr\} \quad \text{subject to} \quad C(f(x')) = 1,
\end{equation}

The author propose a population-based optimization procedure that iteratively applies mutation, selection, and crossover operations on candidate adversarial texts. At each iteration, candidates are modified to improve their overall semantic similarity to x while preserving the misclassification property.

\subsubsection{TextHoaxer}
TextHoaxer addresses this adversarial attack problem in a hard-label black-box setting by formulating the task in the continuous word embedding space. Instead of relying on genetic algorithms that maintain large populations of candidates, TextHoaxer uses a single initialized adversarial candidate and represents the modifications via a perturbation matrix $P$. The optimization objective integrates three key terms: a semantic similarity term $(-\text{sim}(x, x'))$, a pairwise perturbation constraint defined by the sum of squared $L_2$ norms of the per-word perturbation vectors $p_i$, and a sparsity constraint that minimizes the sum of the perturbation magnitudes $|\gamma_i|$.

\begin{equation}
\min_{P} L(P) = \lambda_1 \left( -\operatorname{sim}(x,x') \right) + \lambda_2 \sum_{i=1}^{n} \|p_i\|_2^2 + \lambda_3 \sum_{i=1}^{n} |\gamma_i|, \quad 
\end{equation}

\begin{equation}
\text{subject to} \quad f(x') \neq f(x)
\end{equation}

thereby efficiently generating adversarial examples with high semantic similarity under a tight query budget.

\begin{table*}[ht]
\centering
\caption{Label Noise Attack with different quantities and positions (accuracy). "-" indicates that no specific label candidates were provided in the prompt, thereby rendering the prompt manipulation attack inapplicable.}
\label{tab:comparison}
\resizebox{\textwidth}{!}{%
\begin{tabular}{l *{18}{c}}
\toprule

 & \multicolumn{9}{c}{\textbf{LLaGA}}
 & \multicolumn{9}{c}{\textbf{GraphPrompter}} \\
\cmidrule(lr){2-10}\cmidrule(lr){11-19}

 & \multicolumn{5}{c}{in-domain}
 & \multicolumn{4}{c}{cross-domain}
 & \multicolumn{5}{c}{in-domain}
 & \multicolumn{4}{c}{cross-domain} \\
\cmidrule(lr){2-6}\cmidrule(lr){7-10}\cmidrule(lr){11-15}\cmidrule(lr){16-19}

\textbf{Dataset}
& \makecell{original} & \makecell{50\%\\front} & \makecell{50\%\\after} & \makecell{100\%\\front} & \makecell{100\%\\after}
& \makecell{50\%\\front} & \makecell{50\%\\after} & \makecell{100\%\\front} & \makecell{100\%\\after}
& \makecell{original} & \makecell{50\%\\front} & \makecell{50\%\\after} & \makecell{100\%\\front} & \makecell{100\%\\after}
& \makecell{50\%\\front} & \makecell{50\%\\after} & \makecell{100\%\\front} & \makecell{100\%\\after} \\
\midrule

\textbf{cora}
& 87.82\% & 84.87\% & 88.93\% & 85.42\% & 89.11\%  
& 87.08\% & 88.19\% & 87.27\% & 88.38\%          
& 67.71\% & 58.12\% & 55.54\% & 52.40\% & 48.34\% 
& 64.02\% & 60.52\% & 60.33\% & 59.59\%           
\\

\textbf{pubmed}
& 89.02\% & 85.24\% & 88.06\% & 85.37\% & 88.08\%
& 85.65\% & 87.78\% & 85.52\% & 88.13\%
& 94.35\% & 93.69\% & 93.76\% & 93.05\% & 93.56\%
& 93.66\% & 94.17\% & 93.38\% & 93.99\%
\\

\textbf{OGB-Arxiv}
& 73.20\% & 73.14\% & 71.42\% & 73.12\% & 71.30\%
& 73.34\% & 73.04\% & 72.62\% & 72.40\%
& 72.68\% & - & - & - & -
& - & - & - & -
\\

\textbf{OGB-Products}
& 73.08\% & 72.24\% & 73.26\% & 70.28\% & 72.82\%
& 70.10\% & 73.46\% & 67.30\% & 73.06\%
& 79.98\% & 73.54\% & 80.30\% & 45.92\% & 77.92\%
& 68.84\% & 78.38\% & 34.20\% & 73.74\%
\\

\textbf{Amazon-Computers}
& 87.41\% & 85.03\% & 86.92\%	& 82.64\% & 86.33\%	& 81.24\% & 86.25\% & 77.70\% & 85.61\%
& 78.31\% & 74.46\% & 75.83\% & 67.30\%	& 74.29\% & 41.37\% & 73.08\% & 20.89\% & 69.26\%
\\

\textbf{Amazon-Sports}
&91.32\%	& 85.51\%	&90.06\%	&84.48\%	&89.83\%	&84.90\%	&89.88\%	&82.16\%	&89.07\%
& 91.99\%	&83.49\%	&91.34\%	&80.41\%	&90.77\%	&78.97\%	&90.45\%	&66.86\%	&90.04\%
\\

\bottomrule
\end{tabular}%
}
\end{table*}

\subsection{Graph Structure Attack}
\subsubsection{Nettack}
Nettack is a \emph{local gray-box} graph structure attack on a \emph{surrogate model}. It assumes no direct knowledge of the target model’s architecture or parameters and only requires access to training data. Given a trained surrogate model on the clean graph \(G^{(0)} = (A^{(0)}, X^{(0)})\), Nettack defines the following loss function:
\[
L_s(A, X; W, v_0) = \max_{c \neq c_{\text{old}}} \Bigl([\hat{A}^2 X W]_{v_0, c} - [\hat{A}^2 X W]_{v_0, c_{\text{old}}}\Bigr),
\]
where \(\hat{A}\) is the normalized adjacency matrix, \(W\) is the learned weight matrix of the surrogate GNN, and \(c_{\text{old}}\) is the original class of the target node \(v_0\).

Nettack aims to \emph{increase the log-probabilities of incorrect classes} by modifying features or edges within a specified budget \(\Delta\). For structure attack, we only use it to modify the edges. Formally, it enforces:
\[
\sum_{u} \sum_{i} \bigl| X^{(0)}_{ui} - X'_{ui} \bigr|
\;+\;
\sum_{u < v} \bigl| A^{(0)}_{uv} - A'_{uv} \bigr|
\;\leq\; \Delta,
\]
and restricts changes to a set of attacker nodes \(\mathcal{A}\). The final optimization problem is:
\[
\arg\max_{(A', X') \,\in\, \hat{P}^{G_0}_{\Delta, \mathcal{A}}} \; L_s (A', X'; W, v_0),
\]
where \(\hat{P}^{G_0}_{\Delta, \mathcal{A}}\) is the set of all modified graphs \((A', X')\) respecting the budget constraints. By iteratively choosing edges and features that maximize the surrogate loss, Nettack successfully degrades the classifier’s performance on the target node with minimal perturbations.

\subsubsection{PR-BCD (Projected Randomized Block Coordinate Descent)}
PR-BCD is a \emph{scalable white-box} graph structure attack, which can be \emph{both local and global}. It relaxes the perturbation matrix \(P\) from the discrete set \(\{0,1\}\) to a continuous interval \([0,1]\), enabling gradient-based optimization:
\[
\max_{P} \quad \mathcal{L}\bigl(f_{\theta}(A \oplus P, X)\bigr)
\quad
\text{s.t.}
\quad
\sum P \leq \Delta.
\]

In each iteration, PR-BCD updates only a \emph{randomly sampled block} of edges. Specifically, it modifies the probabilities \(p_t\) of selected edges using the gradient of the loss \(\mathcal{L}(\hat{y}, y)\):
\[
p_t \gets p_{t-1} + \alpha_t \,\nabla_{p_{t-1}[i_{t-1}]} \mathcal{L}(\hat{y}, y),
\]
where \(i_{t-1}\) indexes the sampled edges. A \emph{projection step} ensures that the expected number of flipped edges does not exceed the budget \(\Delta\). To further improve efficiency, PR-BCD periodically \emph{resamples} a portion of the block, discarding edges that have minimal gradient contributions. After training, the final binary perturbation matrix is obtained via Bernoulli sampling:
\[
P \sim \text{Bernoulli}(p_E).
\]

By focusing on sparse, block-wise updates instead of handling the entire adjacency matrix, PR-BCD significantly reduces both time and space complexity, thus enabling large-scale adversarial attacks even on graphs with millions of nodes.

\subsubsection{Defense}
We adopt two adversarial training methods to defend the GraphLLMs against graph structure attacks: FGSM (Fast Gradient Sign Method) and PGD (Projected Gradient Descent).

FGSM minimizes a combination of standard and adversarial losses. Given model parameters \( \theta \), input \( x \), label \( y \), and loss function \( J(\theta, x, y) \), the adversarial training objective is:

\begin{equation}
    \tilde{J}(\theta, x, y) = \alpha J(\theta, x, y) + (1 - \alpha) J(\theta, x + \epsilon \cdot \text{sign} (\nabla_x J(\theta, x, y))).
\end{equation}

where \( \epsilon \) is the perturbation magnitude that controls the strength of the adversarial attack, and \( \nabla_x J(\theta, x, y) \) and \( \alpha \) controls the balance between original and adversarial loss.

PGD iteratively generates stronger adversarial examples compared to FGSM. The adversarial objective is:

\begin{equation}
    \tilde{J}(\theta, x, y) = \alpha J(\theta, x, y) + (1 - \alpha) J(\theta, x_{\text{adv}}),
\end{equation}

where \( x_{\text{adv}} \) is computed iteratively as:

\begin{equation}
    x_{t+1} = \Pi_{x + S} \left( x_t + \epsilon \cdot \text{sign}(\nabla_x J(\theta, x_t, y)) \right)
\end{equation}

where \( \Pi_{x + S} \) is the projection operator, ensuring that the perturbed example remains within a valid perturbation set \( S \), typically an \( \ell_p \)-norm ball around \( x \).

\section{Experiments}
\subsection{GraphLLM Setting}
We employ the experimental configurations of our baseline models as presented in the Tables \ref{tab:graphllmcomp}, \ref{tab:graphllmhyperparam}.
\begin{table}[htbp]
  \centering
  \resizebox{0.5\textwidth}{!}{%
    \begin{tabular}{c c c c}
      \toprule
      Model & Primary embedding & Graph encoder & Base model \\
      \midrule
      LLaGA & Sbert & - & Vicuna-7b \\
      GraphPrompter & GIA & GAT & Llama-2-7b \\
      GraphTranslator & Bert & GraphSAGE & ChatGLM-6b \\
      \bottomrule
    \end{tabular}%
  }
  \caption{GraphLLM Model Components Setting}
  \label{tab:graphllmcomp}
\end{table}
\begin{table}[htbp]
  \centering
  \resizebox{\columnwidth}{!}{%
    \begin{tabular}{c c c c c c c c c}
      \toprule
      Model & Initial LR & Warm-up LR & Scheduler & Batch & Decay & Warm-up Ratio & Epochs & Steps \\
      \midrule
      LLaGA & 2e-3  & -    & Linear Cosine & 16 & 0    & 0.03 & - & - \\
      GraphPrompter & 1e-5 & -    & -                & 12 & 0.05 & -    & 1 & - \\
      GraphTranslator S1 & 1e-4  & 1e-6 & Linear Cosine & 8  & 0.05 & -    & - & 5000 \\
      GraphTranslator S2 & 1e-4  & 1e-6 & Linear Cosine & 1  & 0.05 & -    & - & 5000 \\
      \bottomrule
    \end{tabular}%
  }
  \caption{GraphLLM Model Hyperparameters}
  \label{tab:graphllmhyperparam}
\end{table}
\subsection{Attack-and-Defense methods setting}
\subsubsection{Graph Structure Attack}
For PRBCD, we use the following hyperparameters across all datasets: 
$block_{\text{size}} = 250000$, $lr = 2000$. 
For other hyperparameters, we use the default value in torch\_geometric.

\subsubsection{Adversarial Training}
For FGSM, the hyperparameters for Pubmed, OGB-Products, OGB-Arxiv are: $\epsilon = 1e-3$, $\alpha = 0.8$, for Cora, these are: $\epsilon = 1e-2$, $\alpha = 0.8$. For PGD, we use the following hyperparameters: $\epsilon = 1e-3$, $\alpha = 0.8$, $num_{\text{steps}} = 10,\quad step_{\text{size}} = 2.5 \times 10^{-4}$.

\subsubsection{Representative Prompts Used in Prompt Manipulation Attacks}
\begin{table}[htbp]
\centering
\renewcommand{\arraystretch}{1.1}
\footnotesize
\caption*{Table 12(a): Prompt examples for different attack types on \textbf{LLaGA}.}
\label{tab:prompt-llaga}
\resizebox{\columnwidth}{!}{%
\begin{tabular}{l l p{6.2cm}}  
\toprule
\textbf{Model} & \textbf{Attack Type} & \textbf{Prompt Example} \\
\midrule
\multirow{4}{*}{LLaGA}
& Original & Please predict the most appropriate category for the paper. Choose from the following categories: Case Based, Genetic Algorithms, Neural Networks, Probabilistic Methods, Reinforcement Learning, Rule Learning, Theory. \\
& Label Shuffle & Please predict the most appropriate category for the paper. Choose from the following categories: Genetic Algorithms, Probabilistic Methods, Reinforcement Learning, Neural Networks, Theory, Case Based, Rule Learning. \\
& In-Domain Noise & Please predict the most appropriate category for the paper. Choose from the following categories: Hydrology, cs.GL, Materials Science, Analytical Chemistry, cs.PF, cs.CC, Physical Chemistry, Case Based, Genetic Algorithms, Neural Networks, Probabilistic Methods, Reinforcement Learning, Rule Learning, Theory. \\
& Cross-Domain Noise & Please predict the most appropriate category for the paper. Choose from the following categories: Computer Components, Car Electronics, Electronics, Tennis and Racquet Sports, Russia, Early Learning, Software, Case Based, Genetic Algorithms, Neural Networks, Probabilistic Methods, Reinforcement Learning, Rule Learning, Theory. \\
\bottomrule
\end{tabular}
}
\end{table}

\begin{table}[htbp]
\centering
\renewcommand{\arraystretch}{1.1}
\footnotesize
\caption*{Table 12(b): Prompt examples for different attack types on \textbf{GraphPrompter}.}
\label{tab:prompt-prompter}
\resizebox{\columnwidth}{!}{%
\begin{tabular}{l l p{6.2cm}}
\toprule
\textbf{Model} & \textbf{Attack Type} & \textbf{Prompt Example} \\
\midrule
\multirow{4}{*}{GraphPrompter}
& Original & Please predict the most appropriate category for the paper. Choose from the following categories:\texttt{\string\n}Rule Learning\texttt{\string\n}Neural Networks\texttt{\string\n}Case Based\texttt{\string\n}Genetic Algorithms\texttt{\string\n}Theory\texttt{\string\n}Reinforcement Learning\texttt{\string\n}Probabilistic Methods\texttt{\string\n}Answer: \\
& Label Shuffle & Please predict the most appropriate category for the paper. Choose from the following categories:\texttt{\string\n}Probabilistic Methods\texttt{\string\n}Neural Networks\texttt{\string\n}Theory\texttt{\string\n}Rule Learning\texttt{\string\n}Genetic Algorithms\texttt{\string\n}Case Based\texttt{\string\n}Reinforcement Learning\texttt{\string\n}Answer: \\
& In-Domain Noise & Please predict the most appropriate category for the paper. Choose from the following categories:\texttt{\string\n}Rule Learning\texttt{\string\n}Neural Networks\texttt{\string\n}Case Based\texttt{\string\n}Genetic Algorithms\texttt{\string\n}Theory\texttt{\string\n}Reinforcement Learning\texttt{\string\n}Probabilistic Methods\texttt{\string\n}Hydrology\texttt{\string\n}cs.GL\texttt{\string\n}Materials Science\texttt{\string\n}Analytical Chemistry\texttt{\string\n}cs.PF\texttt{\string\n}cs.CC\texttt{\string\n}Physical Chemistry\texttt{\string\n}Answer: \\
& Cross-Domain Noise & Please predict the most appropriate category for the paper. Choose from the following categories:\texttt{\string\n}Computer Components\texttt{\string\n}Car Electronics\texttt{\string\n}Electronics\texttt{\string\n}Tennis and Racquet Sports\texttt{\string\n}Russia\texttt{\string\n}Early Learning\texttt{\string\n}Software\texttt{\string\n}Rule Learning\texttt{\string\n}Neural Networks\texttt{\string\n}Case Based\texttt{\string\n}Genetic Algorithms\texttt{\string\n}Theory\texttt{\string\n}Reinforcement Learning\texttt{\string\n}Probabilistic Methods\texttt{\string\n}Answer: \\
\bottomrule
\end{tabular}
}
\end{table}

\end{document}